\documentclass[5p]{elsarticle}

\usepackage{geometry}
\usepackage{indentfirst}
\usepackage{amsmath, mathtools}

\usepackage{graphicx, caption}
\usepackage{pifont, amssymb}
\usepackage{multicol}
\usepackage{hyperref}

\usepackage{booktabs, multirow, makecell, diagbox, threeparttable}
\usepackage{color}
\usepackage{stfloats}
\usepackage{float}
\begin{document}
	
	\begin{frontmatter}
		\title{\textbf{First Power Linear Unit with Sign}}
		
		\author{Boxi Duan} \ead{bossey\_dwan@foxmail.com}
		\address{Sun Yat-sen University, Guangzhou 510006, China}
		
		\begin{abstract}
			Convolutional neural networks (CNNs) have shown great dominance in vision tasks because of their tremendous capability to extract features from visual patterns, in 
			which activation units serve as crucial elements for the entire structure, introducing non-linearity to the data processing systematically. This paper proposes a novel 
			and insightful activation method termed FPLUS, which exploits mathematical power function with polar signs in form. It is enlightened by common inverse operation 
			while endowed with an intuitive meaning of bionics. The formulation is derived theoretically under conditions of some prior knowledge and anticipative properties, 
			and then its feasibility is verified through a series of experiments using typical benchmark datasets, whose results indicate our approach owns superior 
			competitiveness among numerous activation functions, as well as compatible stability across many CNN architectures. Furthermore, we extend the function presented to a 
			more generalized type called PFPLUS with two parameters that can be fixed or learnable, so as to augment its expressive capacity, and outcomes of identical tests 
			validate this improvement. Therefore we believe the work in this paper has a certain value of enriching the family of activation units. 
			\newline
		\end{abstract}
	
	\end{frontmatter}

	\section{Introduction}
	\par As a core component embedded in CNN, the activation unit models the synapse in a way of non-linear mapping, transforming different inputs into valid outputs 
	respectively before transmitting to the next node, and this kind of simulation is partly motivated by neurobiology, such as McCulloch and Pitts’ M-P model \cite{M-P_model}, 
	Hodgkin and Huxley’s H-H equation \cite{H-H_equation}, Dayan and Abbott’s firing rate curve, etc. As a consequence, a neural network is inherently non-linear owing to the 
	contribution of hidden layers, in which activation function is the source providing such vitally important property. Thus it can be seen that an appropriate setup of 
	non-linear activation should not be neglected for the neural networks, and the option of this module usually varies according to the scenarios and requirements.
	
	\par Over the years, a variety of activation functions have come into being successively, and some of them attain excellent performance. Sigmoid is applied in the early 
	days but deprecated soon because of the severe gradient vanishing problem, which gets considerable alleviation from ReLU \cite{Boltzmann_Machines, ReLU}. However, the risk of 
	dying neurons makes ReLU vulnerable and hence a lot of variants like Leaky ReLU \cite{Leaky_ReLU} try to fix this shortcoming. Subsequently, ELU \cite{ELU} is proposed to 
	address the bias shift effect while its limitation is relatively settled by supplementary strategies such as SELU \cite{SELU}. Meanwhile, combinative modality is another way 
	to implement that merges different activation functions, e.g. Swish \cite{SiLU, Swish} and Mish \cite{Mish}. As for works including Maxout \cite{Maxout} and ACON 
	\cite{ACON}, they attempt to unify several methods into the same paradigm. Besides, a few ways approximate the sophisticated function through piecewise linear analog for 
	reduction of complexity, like hard-Swish \cite{hard-Swish}.
	
	\par What mentioned above presents a sketchy outline of remarkable progress about activation units in recent decades, that most of them are based on the transform, 
	recombination, or amendment of linear and exponential models, sometimes logarithmic, but rarely seen with mathematical power form, and this arouse the curiosity what if 
	activated by a certain power function rather than those familiar ones? Additionally, few of the methods take bionic characteristics into account and lack neurobiological 
	implication, which is weak to symbolize the mechanism behind a neural synapse.
	
	\par Drawing inspiration from the M-P model \cite{M-P_model} and perceptron theory \cite{Perceptron}, as well as binary connect algorithm \cite{BinaryConnect} and symmetric 
	power activation functions \cite{SymmetricPowerActivation}, this paper proposes a new approach of activation which considers sign function as the switch factor and applies 
	distinct first power functions to positive and negative input respectively, so a bionic meaning is imparted to it in some degree, and we name it as first power 
	linear unit with sign (FPLUS). In terms of what has been discussed in \cite{NoisyActivation} by Bengio et al., we theoretically derive the subtle representation in form of 
	power function under some designated premises, which are in accordance with the attributes that a reasonable activation unit requires. Furthermore, to follow the spirits of 
	PReLU \cite{PReLU}, PELU \cite{PELU}, and other parallel works, we generalize the expression to an adjustable form with two parameters which can be either manually set or 
	learned from the training process, regulating both amplitude and flexure of the function shape, so as to help optimization when fine-tuning, and naturally we call it 
	parametric version (PFPLUS).
	
	\par Of course, we conduct a sequence of experiments to verify the feasibility and robustness of proposed FPLUS and PFPLUS, also compare them with some popular activation 
	units in classification task to examine how our methods perform, which demonstrates that the approach we come up with is capable of achieving comparable and steady results. 
	Meanwhile, in order to figure out a preferable dynamic scope of fluctuation intervals, we explore the setting influence of two controllable parameters by various 
	initialization or assignment, and a preliminary optimization tendency is obtained.
	
	\par The main contributions of our work can be briefly summarized as below: (1) we theoretically propose a new activation formulation FPLUS which contains bionic allusion, and 
	extend it to a generalized form PFPLUS with tunable parameters; (2) we carry out experiments diversely to validate the performance of our methods, and the results demonstrate 
	their advantages of competence; (3) effect of parameters’ variation is explored by tests, which analyses the dynamic property of PFPLUS.
	
	\par The remainder of this paper is divided into five sections as follows. Section \ref{Sec2} reviews some representative activation functions in recent years. Section 
	\ref{Sec3} describes our method's derivation in detail, while its characteristics and attributes are analyzed in section \ref{Sec4}. Section \ref{Sec5} includes 
	experimental results and objective discussion. Eventually, a conclusive summary is presented in section \ref{Sec6}.

	\section{Related Works} \label{Sec2}
	\par
\begin{figure*}[h]
		\centering
		\includegraphics[scale=0.35]{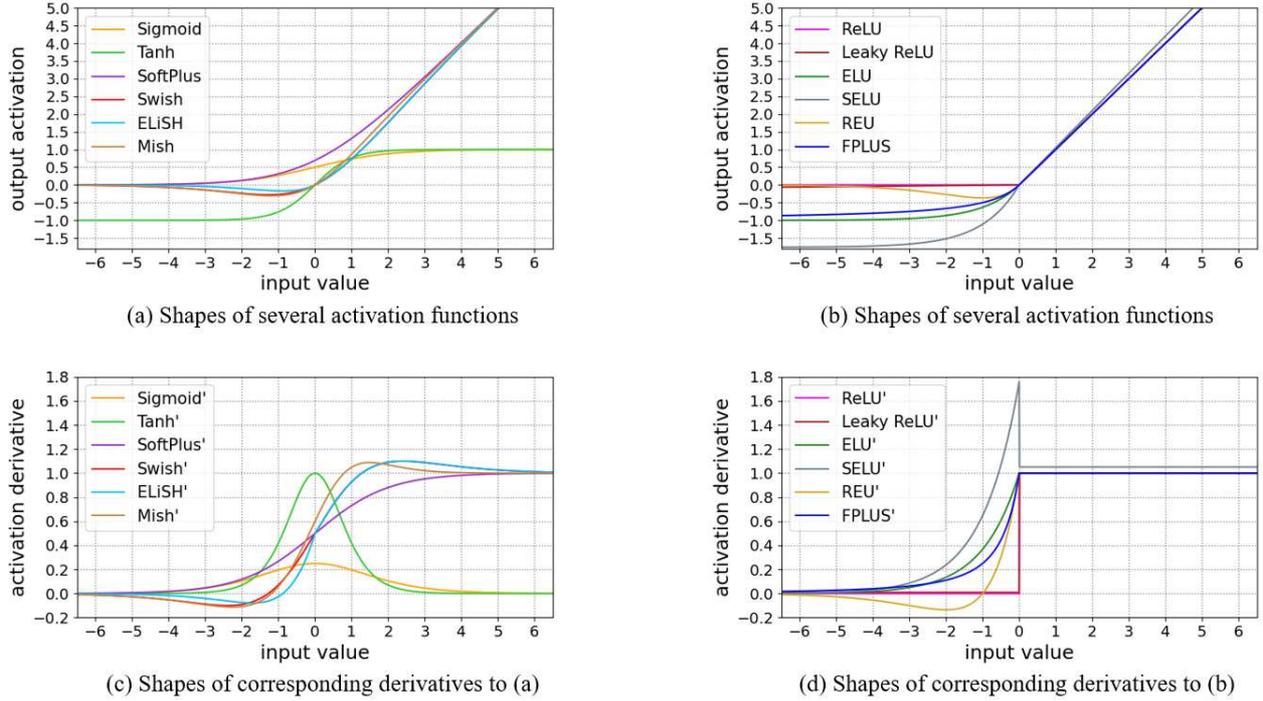}
		\caption{Shapes of typical activation functions and their corresponding derivatives.} 
		\label{Fig.1}
	\end{figure*}

 As is known to all, there exist a number of activation functions so far, some of which obtain significant breakthroughs. We illustrate part of them and our work FPLUS 
	with corresponding derivatives in \autoref{Fig.1}, also we list formulae of the prominent ones below. 
	
	\par Conventional activation function sigmoid is utilized in the early stage, but Xavier showed us it tends to result in gradient vanishing phenomenon 
	\cite{GradientVanishment} for it is bounded in positive domain, so does the tanh function. However, LeCun’s article \cite{TanhFaster} pointed out tanh leads to faster 
	convergence for the network than sigmoid because the former one has zero-mean outputs, which are more in accord with the condition of natural gradient expounded in 
	\cite{NaturalGradient}, and thus iteration can be reduced. Nevertheless, both of them produce saturated output when there comes large positive input, making the parameter's 
	updating probably stagnate.
	
	\begin{equation}
		\mathrm{sigmoid}(x) = \frac{1}{1+e^{-x}} 
	\end{equation}
	
	\begin{equation}
		\mathrm{tanh}(x) = \frac{e^{x} - e^{-x}}{e^{x}+e^{-x}} 
	\end{equation}
	
	\par Afterward, ReLU \cite{Boltzmann_Machines, ReLU} is widely employed as the activation function in CNN. Not only does it have easy shape and concise expression, but also 
	alleviates the problem of gradient diffusion or explosion through identity mapping in positive domain. Still, it forces the output of negative value to be zero, and 
	thereby introduces the distribution of sparsity to the network. Even though this sort of setting seems plausible referring to neurobiological literature \cite{Sparse_Coding, 
	Sparse_Activation}, it as well brings about the disadvantage called dying neurons, which are the units not activated since their gradients reach zero, and in that case, 
	parameters can’t be updated. Therefore to overcome this drawback, Leaky ReLU \cite{Leaky_ReLU} assigns a fixed slope of non-zero constant to the negative quadrant. However, 
	the value specified is extremely sensitive, so PReLU \cite{PReLU} makes it trainable, but this way inclines to cause overfitting. Then RReLU \cite{RReLU} changes it to be 
	stochastically sampled from a probability distribution, resembling random noise so that generating regularization. On the other hand, instead of focusing on the activation 
	function itself, some methods seek assistance from auxiliary features distilled additionally, such as antiphase information for CReLU \cite{CReLU}, and spatial context for 
	FReLU \cite{FReLU}, DY-ReLU \cite{DY-ReLU}, but this means more overhead on pretreatment.
	
	\begin{equation}
		\mathrm{ReLU}(x) = \begin{dcases}
			x, & \text{if } x\geqslant 0 \\
			0, & \text{if } x<0
		\end{dcases}
	\end{equation}
	
	\begin{equation}
		\mathrm{LReLU}(x) = \begin{dcases}
			x, & \text{if } x\geqslant 0 \\
			\alpha x, & \text{if } x<0
		\end{dcases}
		\quad (\alpha > 0)
	\end{equation}
	
	\par In contrast with ReLU, another classic alternative is ELU \cite{ELU}, which replaces the output of negative domain with an exponential function. In this way, ELU 
	avoids the predicament of dead activation from zero derivative, also mitigates the effect of bias shift by pushing the mean of outputs closer to zero. Besides, it appears 
	robust to the noise due to the existence of saturation in negative quadrant. Later CELU \cite{CELU} remedies its flaw of not maintaining differentiable continuously in some 
	cases, while PELU \cite{PELU} gives it more flexibility with learnable parameters. In addition, SELU \cite{SELU} is proved to possess a self-normalizing attribute with respect 
	to the fixed point which is supposed after allocating scale factors, and FELU \cite{FELU} provides a peculiar perspective to realize speeding-up exponential activation by 
	equivalent transformation.
	
	\begin{equation}
		\mathrm{ELU}(x) = \begin{dcases}
			x, & \text{if } x\geqslant 0 \\
			\alpha (e^x-1), & \text{if } x<0
		\end{dcases}
		\quad (\alpha > 0)
	\end{equation}
	
	\begin{equation}
		\mathrm{SELU}(x) = \lambda\cdot\begin{dcases}
			x, & \text{if } x\geqslant 0 \\
			\alpha (e^x-1), & \text{if } x<0
		\end{dcases} \\ 
		(\lambda > 1, \alpha > 0)
	\end{equation}	
	
	\par Besides, the nested or composite mode can be an innovative scheme to generate new activation functions. SiLU \cite{SiLU} for instance, also known as Swish \cite{Swish}, 
	exactly multiplies the input by sigmoidal weight, and ELiSH \cite{ELiSH} makes a little alteration on this basis. GELU \cite{GELU} adopts a similar operation, regarding 
	the cumulative distribution function of Gaussian distribution as the weight, while Mish \cite{Mish} combines tanh and SoftPlus \cite{Softplus} before weighting the 
	input. As for REU \cite{REU}, it directly takes an exponential function to be the multiplier which looks like a hybrid of ReLU and ELU, whereas RSigELU \cite{RSigELU} even 
	involves extra sigmoid at the cost of making the expression more complicated.
	
	\begin{equation}
		\mathrm{Swish}(x) = x \cdot sigmoid(\beta x) \quad (\beta > 0)  
	\end{equation}	
	
	\begin{equation}
		\mathrm{GELU}(x) = x \cdot \frac{1}{\sqrt{2\pi}} \int_{-\infty}^x e^{-\frac{t^2}{2}} \mathrm{d}t
	\end{equation}		
	
	\begin{equation}
		\mathrm{Mish}(x) = x \cdot tanh[ln(1+e^x)]
	\end{equation}

	\par It is likewise noticed that piecewise linear approximation of a curve is usually requested in the mobile devices restricted by computing resources, and thus hard-series 
	fitting was put forward, such as hard-sigmoid \cite{BinaryConnect}, hard-Swish \cite{hard-Swish}. Moreover, some other works like Maxout \cite{Maxout}, ACON \cite{ACON} 
	managed to unify multiple types of activation methods into one single paradigm so that each of them becomes a special case to the whole family, yet this means much more 
	parameters have to be learned to fulfill such kind of adaptive conversion, and hence preprocessing efficiency might be undesired.
	
	\section{Proposed Method} \label{Sec3}
	In this section, we first explain the motivation of our approach more specifically, and the deduction process is described concretely in the next paragraph.
	\subsection{\textsf{Source of Motivation}}
	\par There is no doubt neurobiology offers probative bionic fundamentals to the design of an imitated neural system, for example, CNN is such a milestone that greatly 
	promotes the advancement of computer vision, and it can be traced back to Hubel and Wiesel’s work \cite{CNN_Inspiration}, whose discovery is an inspiration for the conception 
	of CNN framework. Hence bionics-oriented practice sometimes might be a way to find clues or hints of creation \cite{BRU}, to which lots of documents attach 
	great importance.
	
	\par M-P model \cite{M-P_model} follows the principle of a biological neuron, that there exist excitatory and inhibitory two states depending on the polarity of input type, 
	and a threshold determines whether it is activated or not for the neuron. Based on this essence, primitive perceptron \cite{Perceptron} utilizes sign function to produce a 
	binary output of type 0 or 1, representing different activation states, so as to abstract out effective and notable information from features generated by lower-level and pass 
	it on to the deep, albeit this is a roughly primeval means. Another recent research called binary connect \cite{BinaryConnect} inherits the idea alike, not that to cope with 
	output but binarize the weight via 1 and -1, whose purpose is decreasing resource consumption on mobile facilities, and such setup could be reckoned as a similar strategy of 
	regularization.
	
	\par On the other hand, advancement has witnessed multiple activation methods come into being and evolve consecutively, and plenty of mathematical functions have been 
	employed to act as transfer units in neural networks. Still, in comparison to piecewise linear, exponential, and logarithmic formulae, power type functions seem not much 
	popular with little attention, so they are rarely seen in the field. Under the impact of those famous pioneers, only a few works focus on this sort, e.g., \cite{PFLU} creates 
	a power-based composite function, but appears tangled because it introduces fraction and square root at the same time, while \cite{SymmetricPowerActivation} changes the 
	positive representation of ReLU by using a couple of power function and its inverse, which are symmetric to the linear part of ReLU, but it lacks further validation on more 
	datasets and networks.
	
	\par As a consequence, there are many potentialities we can explore and exploit in power type activation theory, and our work probes into how to construct a refined activation 
	function of this kind with simplicity as well as effectiveness, and pays more attention to the contact with bionic meanings.
	
	\subsection{\textsf{Derivation Process}}
	\par Obviously, the positive part $ y=x $ of ReLU is identical mapping, however, it can be also regarded as $ y=x^1$, i.e. the output is one power over the input, and 
	this is for positive situation, which reflects the excitatory state of a neuron. Associating with implicit bionic allusion, we ponder the question can we represent the 
	inhibitory state in a contrary way based on a certain power type function, corresponding to the negative part of the same united expression through inverse operation?
	
	\par As the reverse procedure of positive first power, negative first power is doubtless supposed to be allocated to the opposite position in the formula, which will involve 1 
	and	-1 two constant exponents simultaneously in that case, so a sign function is taken into consideration to play the role of switching between them. Meanwhile, the 
	first-order power is relatively common and simple than those higher-order ones, consuming less computational cost while enabling itself to handle bilateral inputs via two 
	distinct branches.
	
	\par For more generality, we firstly bring in some coefficients $\alpha$, $\beta$, $\omega$, $\theta$ which are finite real numbers and unequal to zero. Besides we give the 
	definition of our formula $ f(x) = \alpha[\omega x + \beta]^{sgn(x)} + \theta $, where sgn$ (x) $ is the sign function shown as Eq.\eqref{SignFunction}, while a 
	piecewise form of our expression is given in Eq.\eqref{PiecewiseDefinition}.

	\begin{equation}
		\mathrm{sgn}(x) = \begin{dcases}
			1, & \text{if } x \geqslant 0 \\
			-1, & \text{if } x < 0
		\end{dcases}
		\label{SignFunction}
	\end{equation}

	\begin{equation}
		f(x) = \begin{dcases}
			\alpha \omega x + \alpha \beta + \theta, & \text{if } x \geqslant 0 \\
			\frac{\alpha}{\omega x + \beta} + \theta, & \text{if } x < 0
		\end{dcases}
		\label{PiecewiseDefinition}
	\end{equation}

	\par With reference to the usual characteristics and prior knowledge of those typical activation methods, we design our function $ f(x) $ to have such expected properties 
	enumerated as follows:\\\\
	\uppercase\expandafter{\romannumeral1}. \ \ Defined on the set of real numbers: \[ \left. x \right |_{f(x)=\infty} > 0 \]
	\uppercase\expandafter{\romannumeral2}. \ Through the origin of coordinate system: \[ f(0) = 0 \]
	\uppercase\expandafter{\romannumeral3}. Continuous at the demarcation point: \[ \lim\limits_{x \to 0^{+}} f(x) = \lim\limits_{x \to 0^{-}} f(x) = f(0) \]
	\uppercase\expandafter{\romannumeral4}. Differentiable at the demarcation point: \[ \left. \dfrac {\partial f(x)}{\partial x} \right|_{x=0^{+}} = \left. \dfrac {\partial 
		f(x)}{\partial x} \right|_{x=0^{-}} \]
	\uppercase\expandafter{\romannumeral5}. \ Monotone increasing: \[ \forall x \in \mathbb{R}  \ , \  \dfrac {\partial f(x)}{\partial x} \geqslant 0 \]
	\uppercase\expandafter{\romannumeral6}. Convex down: \[ \forall x \in \mathbb{R}  \ , \   \dfrac {\partial^2 f(x)}{\partial x^2} \geqslant 0 \]
	
	\par Condition \uppercase\expandafter{\romannumeral1} requires any value on the set of real numbers ought to be valid as input, but we see there exists trouble when the 
	negative part of $ f(x) $ expression achieves infinity since the denominator $ \omega x + \beta $ of the fraction isn't allowed to be zero. As a consequence, considering the 
	piecewise attribute, we can just confine the pole point to the opposite interval and this latent defect can be forever avoided. Assuming that $ \exists x_i \in (-\infty,0) 
	$ , which makes for 
	\[ \omega x_i + \beta = 0  \implies x_i = - \dfrac{\beta}{\omega} \]  
	To cause a paradox, we need 
	\[ x_i > 0 \implies \dfrac{\beta}{\omega} < 0 \ \cdot \cdot \cdot \cdot \ \text{\ding{172}} \]  
	
	\par With regard to condition \uppercase\expandafter{\romannumeral2}, it is clear that point $ (0,0) $ of the orthogonal plane coordinate system must be a solution to our 
	activation function, and thus we have 
	\[ f(0) = 0 \implies \alpha \beta + \theta = 0 \ \cdot \cdot \cdot \cdot \ \text{\ding{173}} \]

	\par Condition \uppercase\expandafter{\romannumeral3} calls for continuity, so limits on both sides of the demarcation point must exist, while they equal each other and 
	also the value of that point. Hence we list
	\[\left. \begin{gathered}
		\lim\limits_{x \to 0^{+}} f(x) = \alpha \beta + \theta \\
		\lim\limits_{x \to 0^{-}} f(x) = \frac{\alpha}{\beta} + \theta
	\end{gathered} \right \}
	\implies \alpha \beta + \theta = \dfrac{\alpha}{\beta} + \theta \]
	Because $ \alpha , \beta \neq 0 $ stated above, it comes out 
	\[ \beta = \pm 1 \ \cdot \cdot \cdot \cdot \ \text{\ding{174}} \]  
	From formula \ding{173} it is known that $ f(0)=0 $ , and solving these simultaneous equations we can obtain 
	\[ \alpha = \begin{dcases}
			-\theta, & \text{if } \beta = 1 \\
			\theta, & \text{if } \beta = -1
		\end{dcases} \ \cdot \cdot \cdot \cdot \ \text{\ding{175}} \] 

	\par As for condition \uppercase\expandafter{\romannumeral4}, it demands taking the derivatives on both sides of the demarcation point and they should be identical, too. 
	Thereby we get
	\[\left. \alpha \omega \right|_{x=0^{+}} = \left. \dfrac {- \alpha \omega}{(\omega x + \beta)^2} \right|_{x=0^{-}} \implies \alpha \omega = \dfrac{-\alpha \omega}{\beta^2} \]
	Since $ \alpha, \beta, \omega \neq 0 $ as previously mentioned, it turns out 
	\[ \beta = i \]
	Now that an outcome of imaginary number $ i $ occurs, which disobeys the hypothesis that $\beta$ is a real number, the precondition thus needs to be amended. 
	
	\begin{figure}[!tp]
		\centering
		\includegraphics[scale=0.2]{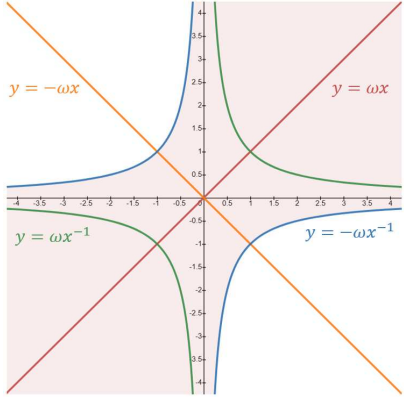}
		\caption{Graphic curves of power functions in a pair of opposite signs with 1 and -1 being exponents. Shadow region signifies a general orientation for $ y=\omega x $ to 
		reach tangency with $ y=-\omega x^{-1} $.} 
		\label{Fig.2}
	\end{figure}
	
	\par Observing the graphic curves of a system for the first power functions, sketchily shown as \autoref{Fig.2} for example, it can be recognized that line $ y=\omega x $ is 
	capable of being tangent to curve $ y=-\omega x^{-1} $ somehow, but that will never happen to curve $ y=\omega x^{-1} $ because they are always of intersection, and vice versa 
	for line $ y=-\omega x $. This fact indicates that the former combination has the potentiality to construct a smooth and derivable function if splicing appropriate portions 
	severally, whereas the latter one does not. The shadow region in the figure implies a possible scope for line $ y=\omega x $ to achieve tangency with curve $ y=-\omega x^{-1} 
	$, while we can notice it crosses curve $ y=\omega x^{-1} $ under any circumstances.
	
	\par Based on the rules mentioned above, we are aware that for reciprocal types like $ y=\omega_1 x $ and $ y=\omega_2 x^{-1} $, only if the coefficients $\omega_1$ and 
	$\omega_2$ have contrary signs, can they be differentiable at the demarcation point in piecewise combination when carrying out fundamental functional transformation, such as 
	flip, scaling, translation, etc. Therefore we know
	\[ \omega_1 \neq \omega_2 \]
	and the original definition of our function has to be adjusted to 
	\[
	f(x) = \begin{dcases}
		\alpha \omega_1 x + \alpha \beta + \theta, & \text{if } x \geqslant 0 \\
		\frac{\alpha}{\omega_2 x + \beta} + \theta, & \text{if } x < 0
	\end{dcases}
	\]
	where $ \alpha, \beta, \omega_1, \omega_2, \theta \neq 0 $ and all of them are finite real numbers.
	In this way, we evaluate the differentiability of the boundary point again
	\[
	\left. \alpha \omega_1 \right|_{x=0^{+}} = \left. \dfrac {- \alpha \omega_2}{(\omega_2 x + \beta)^2} \right|_{x=0^{-}} \implies \omega_1 = \dfrac{-\omega_2}{\beta^2} 
	\]
	Substituting formula \ding{174} into the equation, we are able to figure out the qualitative and quantitative relation
	\[ \omega_1 = -\omega_2 \ \cdot \cdot \cdot \cdot \ \text{\ding{176}} \]
	In the meantime, inequality \ding{172} is hence changed into
	\[ \dfrac{\beta}{\omega_2} < 0 \]
	which can be inferred that
	\[\omega_2 \begin{dcases}
		<0, & \text{if } \beta = 1 \\
		>0, & \text{if } \beta = -1 
	\end{dcases} \ \cdot \cdot \cdot \cdot \ \text{\ding{177}} \]

	\par As far as condition \uppercase\expandafter{\romannumeral5} is concerned, it involves property of first-order derivative, and owing to the premise that every coefficient 
	is never zero, we accordingly expect \\\\
	(\romannumeral1) $ \ \forall x \geqslant 0, \ \dfrac{\partial f(x)}{\partial x} = \alpha \omega_1 > 0  \ \cdot \cdot \cdot \cdot \ \text{\ding{178}} $ \\\\
	(\romannumeral2) $ \ \forall x < 0, \ \dfrac{\partial f(x)}{\partial x} = \dfrac{-\alpha \omega_2}{(\omega_2 x + \beta)^2} > 0 \ \implies \alpha \omega_2 < 0 $ \\\\
	Consulting formula {\ding{176}}, it is obvious the above two restrictions are equivalent and meeting one of them e.g. inequality {\ding{178}} is quite enough.
	
	\par Finally, it comes to condition \uppercase\expandafter{\romannumeral6}, which asks for the function's being convex down and that refers to second-order derivative, thus we 
	hold 
	\\\\
	(\romannumeral1) $\forall x \geqslant 0, \ \dfrac{\partial^2 f(x)}{\partial x^2} = 0 \geqslant 0  $ \\\\
	(\romannumeral2) $\forall x < 0, \ \dfrac{\partial^2 f(x)}{\partial x^2} = \dfrac{2\alpha \omega_2^2}{(\omega_2 x + \beta)^3} > 0 \implies \dfrac{\alpha}{\omega_2 x + \beta} > 
	0 $ \\\\
	Subordinate condition (\romannumeral1) is apparently self-evident, while for (\romannumeral2) it can be found out from the formulae \ding{176} $\thicksim$ \ding{178} that
	\begin{enumerate}[$\rhd$]
		\item when $\beta=1$, then $ \omega_2<0, \ \alpha>0 \implies \forall x \in (-\infty, 0)$ subordinate condition (\romannumeral2) is tenable
		\item when $\beta=-1$, then $ \omega_2>0, \ \alpha<0 \implies \forall x \in (-\infty, 0)$ subordinate condition (\romannumeral2) is tenable
	\end{enumerate}
	As a result, it means condition \uppercase\expandafter{\romannumeral6} is inevitable on the basis of previous constraints.
	
	\par To sum up, any group of real numbers $\alpha, \beta, \omega_1, \omega_2, \theta$ which are not equal to zero and satisfy formulae \ding{174} $\thicksim$ \ding{178}, 
	can be valid  coefficients for our function, and we summarize them as two cases.
	\begin{enumerate}[{$\bullet$ case} 1:]
		\item $\beta=1, \ \omega_2<0, \ \omega_1=-\omega_2>0, \ \alpha>0, \ \theta=-\alpha$
		\item $\beta=-1, \ \omega_2>0, \ \omega_1=-\omega_2<0, \ \alpha<0, \ \theta=\alpha$
	\end{enumerate}

	\par Having case 1 and case 2 substituted into the latest definition of $f(x)$ separately and simplifying them, hence there comes
	\[\dagger \text{ case 1: if }\beta=1, \ \text{then }f(x)=\alpha \omega_1 \cdot \begin{dcases}
			\ x, & \text{if } x \geqslant 0 \\
			\dfrac{x}{1-\omega_1 x}, & \text{if } x < 0 
		\end{dcases}\] 
	in which $\alpha>0, \ \omega_1 > 0$;
	\[\dagger \text{ case 2: if }\beta=-1, \ \text{then }f(x)=\alpha \omega_1 \cdot \begin{dcases}
		\ x, & \text{if } x \geqslant 0 \\
		\dfrac{x}{1+\omega_1 x}, & \text{if } x < 0 
	\end{dcases}\] 
	in which $\alpha<0, \ \omega_1<0$.
	
	\par Observing the two situations acquired above, it's not hard to notice that $ \alpha \omega_1 > 0 $ always stands, and the coefficient of x laying in the denominator of 
	$f(x)$'s piecewise representation within $(-\infty, 0)$, is always true to be negative. Therefore we can unify the two cases that $ \beta=\pm 1 $ for the following new form:
	\begin{equation}
		f(x)=\lambda \cdot \begin{dcases}
			\ x, & \text{if } x \geq 0 \\
			\dfrac{x}{1-\mu x}, & \text{if } x < 0
		\end{dcases} \quad (\lambda>0,\mu>0)
		\label{PFPLUS-Piecewise}
	\end{equation}
	
	\par Here $\lambda$ is termed as amplitude factor, which controls the degree of saturation for negative zone and the magnitude of slope for positive zone, while $\mu$ is 
	called scale factor, which regulates the speed of attenuation for negative domain only. The derivative of our formulation is given below:
	\begin{equation}
		f^\prime (x)=\lambda \cdot \begin{dcases}
			\ 1, & \text{if } x \geqslant 0 \\
			\dfrac{1}{(1-\mu x)^2}, & \text{if } x < 0
		\end{dcases} \quad (\lambda>0,\mu>0)
		\label{PFPLUS_Derivative-Piecewise}
	\end{equation}
	
	\par So far it can be seen that Eq.\eqref{PFPLUS-Piecewise} is probably the core of our theory, which pretty resembles SELU\cite{SELU} in conception when taken as a 
	counterpart since they both have linear and decayed areas with two controlled parameters that facilitate their shifting in shapes.

	\section{Analysis of Characteristics and Attributes} \label{Sec4}
	\par After demonstrating the deduction process of our means and giving the ultimate definition in segmented form, in this section we step further to dig into more traits and 
	properties of it.
	
	\par In order to merge piecewise activation into one united paradigm for the response, therefore we introduce the Heaviside step function known as 
	Eq.\eqref{HeavisideStepFunction}, i.e. unit step function, so that enable it to classify the polarity of input's sign, and react like a switch in neural circuit, transferring 
	the signal with instant gain or gradual suppression. 
	\begin{equation}
		H(x)=\begin{dcases}
			1, & \text{if } x \geqslant 0 \\
			0, & \text{if } x < 0
		\end{dcases}
		\label{HeavisideStepFunction}
	\end{equation}
	
	\par As long as applying conversion to Eq.\eqref{PFPLUS-Piecewise}, we hereby formally set forth parametric first power linear unit with sign (PFPLUS):
	\begin{equation}
		\mathrm{PFPLUS}(x) = \lambda x \cdot (1-\mu x)^{H(x)-1}
		\label{PFPLUS}
	\end{equation}
	where $H(x)$ is Heaviside step function, and $ \lambda, \mu>0 $.
	
	\par
\begin{figure}[!t]
		\centering
		\includegraphics[scale=0.2]{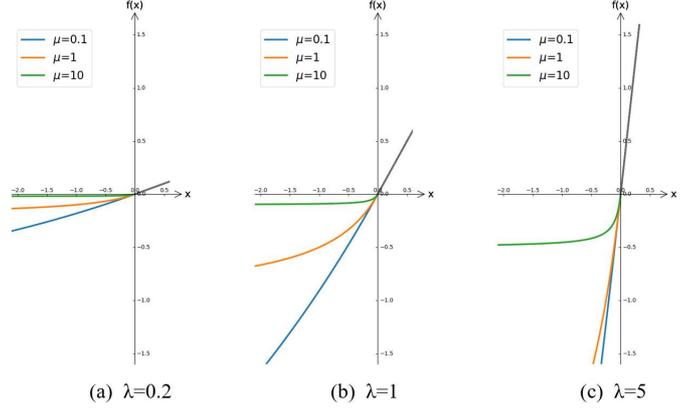}
		\caption{Different shapes of PFPLUS when $\lambda$ and $\mu$ vary.} 
		\label{Fig.3}
	\end{figure}

 Undisputedly, with parameters $\lambda$ and $\mu$ changing, our function proposed will present different shapes in the graph. As shown in \autoref{Fig.3}, it is similar 
	to ReLU when $\lambda=0.2$ and $\mu=10$, while approximates linear mapping as $\lambda=5$ and $\mu=0.1$. Of course, they can be either learned from training procedure or fixed 
	by initial setting, depending on distinct purpose.
	
	\par About trainable PFPLUS in the implementation, we choose established gradient descent as the iterative approach to updating and learning these two coefficients. More than 
	that, we adopt strategies like PReLU \cite{PReLU} to avoid overfitting as much as possible. For every activation layer of PFPLUS, parameters $\lambda$ and $\mu$ are completely 
	identical across all of the channels, in other words, they are channel-shared or channel-wise. As a result for the whole network, the increment of parameters is merely 
	negligible to the total quantity of weights. 
	
	\par Beyond that, Ramachandran et al. have done a lot of work searching for good activation units based on ReLU \cite{Swish_Self-Gated}, including unary functions and binary 
	functions. They discovered that those which perform well tend to be concise, often consisting of no more than two elements, and they usually comprise original input, i.e. in 
	the form of $ f[x, g(x)]$. The study also perceived that most of them are smooth and equipped with linear regions. However, they missed inquiring into power ones with 
	negative exponents when researching on unary types, whereas our work makes some supplementation in a sense.
	
	\par Particularly, as the parameters $\lambda$ and $\mu$ are both equal to one, the function becomes a special form like Eq.\eqref{FPLUS_piecewise}, which is the simplest 
	pattern of its whole family. In addition, considering the views Bengio et al. discussed in \cite{NoisyActivation}, we theoretically reckon the advantages for our way as below:
	
	\begin{enumerate}[1.]
		\item For positive zone, identity mapping is retained for the sake of avoiding gradient vanishing problem.
		\item For negative zone, as the negative input goes deeper, the outcome will appear a tendency to gradually reach saturation, which offers our method robustness to the 
		noise. 
		\item Entire outputs' mean of the unit is close to zero, since the result yielded for negative input isn't directly zero,  but exists a relative minus response to 
		neutralize the holistic activation, so that bias shift effect can be reduced.
		\item When having the corresponding formulation of negative part processed in Taylor expansion, seen as Eq.\eqref{Taylor_expansion} , the operation carried out in negative 
		domain is equivalent to dissociating each order component of the input signal received, and thus more abundant features might be attained up to a point.
		\item From an overall perspective, the shape of function is unilateral inhibitive, and this kind of one-sided form could facilitate the introduction of sparsity to output 
		nodes, making them be like logical neurons.
	\end{enumerate}
	
	\begin{equation}
		f(x) = \begin{dcases}
			\ x, & \text{if } x \geqslant 0 \\
			\dfrac{x}{1-x}, & \text{if } x < 0
		\end{dcases}
		\label{FPLUS_piecewise}
	\end{equation}
	
	\begin{equation}
			\dfrac{x}{1-x} = x + x^2 + x^3 + \dots + x^{n-1} + x^n + O(x^n)
		\label{Taylor_expansion}
	\end{equation}
	
	Moreover, our method can be further associated with a bionic meaning, as long as we implant sign function listed in Eq.\eqref{SignFunction} to play the role of neuron switch, 
	and therefrom we officially put forward first power linear unit with sign (FPLUS), shown as Eq.\eqref{FPLUS_sgn}. 
	
	\begin{equation}
		\mathrm{FPLUS}(x) = [\mathrm{sgn}(x) \cdot x + 1]^{\mathrm{sgn}(x)} - 1 
		\label{FPLUS_sgn}
	\end{equation}
	
	\par For the two sign functions applied in this definition, the one outside bracket is an exponent for the whole power, and it can be regarded as a discriminant factor to the 
	input's polarity, while the one inside bracket is a coefficient, which can be seen as a weight to the input, working in a bipolar way somewhat like \cite{BinaryConnect} 
	mentioned.

	\par In conclusion, what expounded above demonstrates our method is in possession of rigorous rationality and adequate novelty. At the same time, there are still numerous 
	other manners we can develop to comprehend its ample connotation.
	
	\section{Experiments and Discussion} \label{Sec5}
	\par In this section, we conduct a series of experiments to probe into feasibility, capability, and universality of our method, also validate those inferences we discussed 
	before by classification task, covering controlled trials and ablation study. In order to explore its influence on the performance of extracting features in neural networks, 
	we test it on different CNN architectures across many representative datasets, including MNIST \& Fashion-MNIST, Kaggle's Dogs Vs Cats, Intel Image Classification, and 
	CIFAR-10.
	
	\par This section is organized by different experiments with each dataset listed above, so it's divided into 5 main segments.

	\subsection{\textsf{Experiments on MNIST \& Fashion-MNIST}}
	\par MNIST\footnote{URL: http://yann.lecun.com/exdb/mnist/} is an old but classic dataset created by LeCun et al. in the 1990s, which is made up of handwritten digits $0\sim9$ 
	with a training set of 60K examples and a test set of 10K examples. These morphologically diverse digits have been self-normalized and centered in grayscale images with 
	$28\times28$ pixels for each. Similarly, Fashion-MNIST\footnote{URL: https://github.com/zalandoresearch/fashion-mnist} is a popular replacement that emerged in recent years, 
	and it shares the same image size, sample quantity, as well as structure of training and testing splits just like the original MNIST. As the name implies, this dataset is a 
	version of fashion and garments, containing labels from 10 classes.
	
	\par Seeing there are two mutable factors in our parametric function, we plan to preliminarily explore their effects on the performance of activation as the values vary. To 
	facilitate direct observation of dynamic change and prominent differences, we choose LeNet-5 \cite{LeNet-5} to learn from the data due to its simplicity and origin. 
	
	\par What comes to the first is an experiment with fixed configurations of $\lambda$ and $\mu$, set with four orders of magnitude for each, which are 0.01, 0.1, 1, and 10. We 
	train the model with MNIST and Fashion-MNIST for 5 epochs separately but follow the same training setups. The batch size is 64 and the learning rate is 0.001, with 
	cross-entropy loss being the loss function and Adam being the optimizer.

	\begin{table*}[!ht]
		\centering
		\caption{Training loss and test accuracy on MNIST \& Fashion-MNIST.}
		\small
		\begin{threeparttable}
			\begin{tabular}{clcccccccc}
				\toprule
				\multirow{2}*{Dataset}		 &Factor & \multicolumn{4}{c}{Loss}      & \multicolumn{4}{c}{Accuracy} \\
				\cmidrule{2-2} \cmidrule(lr){3-6} \cmidrule(lr){7-10}
				& \diagbox{$\lambda$}{$\mu$} 		 & 0.01  & 0.1 	 & 1 	 & 10 	 & 0.01    & 0.1 	 & 1 	   & 10 \\
				\midrule
				\multirow{4}*{MNIST} 		 & 0.01  & 1.573 & 1.485 & 1.441 & 1.534 & 39.93\% & 43.85\% & 44.89\% & 41.20\% \\
				& 0.1   & 0.147 & 0.114 & 0.139 & 0.158 & 96.21\% & 97.01\% & 96.63\% & 96.05\% \\
				& 1     & 0.054 & 0.046 & \color{blue}0.027 & 0.028 & 98.40\% & 98.31\% & \color{blue}98.97\% & 98.92\% \\
				& 10    & 0.567 & 0.430 & 0.111 & 0.079 & 96.21\% & 96.47\% & 97.11\% & 97.45\% \\
				\midrule
				\multirow{4}*{Fashion-MNIST} & 0.01  & 1.091 & 1.072 & 1.065 & 1.343 & 52.88\% & 57.25\% & 59.45\% & 52.30\% \\
				& 0.1   & 0.630 & 0.568 & 0.534 & 0.579 & 76.49\% & 78.84\% & 79.87\% & 77.97\% \\
				& 1     & 0.309 & 0.307 & \color{blue}0.253 & 0.267 & 87.75\% & 88.18\% & \color{blue}89.62\% & 88.98\% \\
				& 10    & 0.496 & 0.636 & 0.371 & 0.341 & 83.49\% & 84.99\% & 86.53\% & 86.71\% \\
				\bottomrule
			\end{tabular}
			\begin{tablenotes}
				\footnotesize
				\item $\circledcirc$ For each dataset, the best results of loss and accuracy are denoted in \textcolor{blue}{blue} color.
			\end{tablenotes}
		\end{threeparttable}
		\label{Table1}
	\end{table*}
	
	\begin{table*}[!t]
			\centering
			\caption{Parameters' adaptive variation for each activation layer after trained on MNIST.}
			\small
			\begin{tabular}{ccccccccccc}
				\toprule
				\multirow{2}*{Initial} & \multirow{3}*{\makecell{Training \\ Loss }} & \multirow{3}*{\makecell{Test \\ Accuracy}} & 
				\multicolumn{8}{c}{Learnable Parameters for Each Activation Layer} \\
				\cmidrule{4-11}
				 &       &       & \multicolumn{2}{c}{Activation1} & \multicolumn{2}{c}{Activation2} & \multicolumn{2}{c}{Activation3} & \multicolumn{2}{c}{Activation4} \\
				 \cmidrule{1-1} \cmidrule(lr){4-5} \cmidrule(lr){6-7} \cmidrule(lr){8-9} \cmidrule(lr){10-11}
				 $\lambda$ and $\mu$  & 	  &  &$\lambda_1$&$\mu_1$ &$\lambda_2$&$\mu_2$ &$\lambda_3$&$\mu_3$ &$\lambda_4$&$\mu_4$ \\
				 \midrule
				 $\lambda, \mu = 0.2$ & 0.003 & 98.90\% & 0.8294 & 0.1312 & 0.8121 & 0.4096 & 0.4798 & 0.7393 & 0.3379 & 0.6136 \\
				 $\lambda, \mu = 0.5$ & 0.003 & 99.08\% & 1.0320 & 0.7777 & 0.8125 & 0.8921 & 0.5349 & 1.1050 & 0.4464 & 0.7673 \\
				 $\lambda, \mu = 1	$ & 0.002 & 99.17\% & 1.1902 & 1.2234 & 1.0173 & 1.2795 & 0.7196 & 1.5806 & 0.6738 & 1.2830 \\
				 $\lambda, \mu = 2	$ & 0.004 & 99.00\% & 1.5427 & 2.2394 & 1.0349 & 2.4056 & 0.8685 & 2.4924 & 0.8976 & 2.0421 \\
				 $\lambda, \mu = 5	$ & 0.005 & 98.99\% & 1.8366 & 6.0857 & 1.1883 & 5.5873 & 1.1241 & 5.6929 & 1.0646 & 5.9816 \\
				 \bottomrule 
			\end{tabular}
			\label{Table2}
	\end{table*}
	
	\begin{table*}[!t]
		\centering
		\caption{Parameters' adaptive variation for each activation layer after trained on Fashion-MNIST.}
		\small
		\begin{tabular}{ccccccccccc}
			\toprule
			\multirow{2}*{Initial} & \multirow{3}*{\makecell{Training \\ Loss }} & \multirow{3}*{\makecell{Test \\ Accuracy}} & 
			\multicolumn{8}{c}{Learnable Parameters for Each Activation Layer} \\
			\cmidrule{4-11}
			&       &       & \multicolumn{2}{c}{Activation1} & \multicolumn{2}{c}{Activation2} & \multicolumn{2}{c}{Activation3} & \multicolumn{2}{c}{Activation4} \\
			\cmidrule{1-1} \cmidrule(lr){4-5} \cmidrule(lr){6-7} \cmidrule(lr){8-9} \cmidrule(lr){10-11}
			$\lambda$ and $\mu$  & 	  &  &$\lambda_1$&$\mu_1$ &$\lambda_2$&$\mu_2$ &$\lambda_3$&$\mu_3$ &$\lambda_4$&$\mu_4$ \\
			\midrule
			$\lambda, \mu = 0.2$ & 0.109 & 90.18\% & 1.3346 & 0.0297 & 1.0644 & 0.6138 & 0.7340 & 0.8240 & 0.3481 & 1.1056 \\
			$\lambda, \mu = 0.5$ & 0.063 & 90.28\% & 1.6568 & 1.0075 & 1.4744 & 1.0531 & 0.9257 & 1.5719 & 0.5796 & 1.5243 \\
			$\lambda, \mu = 1  $ & 0.044 & 90.36\% & 1.6525 & 1.6226 & 1.7972 & 1.1518 & 1.1746 & 2.2956 & 0.8508 & 2.0996 \\
			$\lambda, \mu = 2  $ & 0.037 & 90.02\% & 1.8306 & 3.0781 & 2.1173 & 1.6818 & 1.4963 & 3.3525 & 1.2920 & 2.8767 \\
			$\lambda, \mu = 5  $ & 0.048 & 89.49\% & 2.7184 & 6.6489 & 2.5286 & 4.5443 & 2.3847 & 6.0374 & 2.3395 & 5.4461 \\
			\bottomrule 
		\end{tabular}
		\label{Table3}
	\end{table*}
	
	\par The results of training loss and test accuracy on both datasets are listed in \autoref{Table1}, grouped by different settings of $\lambda$ and $\mu$. Items in blue 
	color are the best outcomes of loss and accuracy relevant to the corresponding configuration of $\lambda$ and $\mu$, which can be seen that our method achieves superior result 
	when the two parameters are equal to 1, getting the lowest loss and highest accuracy than any other combinations for both MNIST and Fashion-MNIST. In addition, factor 
	$\lambda$ seems to be more sensitive than $\mu$ and exhibits a more intense response as the value changes. For each dataset, only considering $\lambda$ with the same $\mu$, 
	results will get better when $\lambda$ grows from 0.01, especially by a large margin for the interval of 0.01 to 0.1, and reaches the optimum around 1, then begins to fall 
	back a little bit. A similar situation happens to $\mu$ basically, but not that conspicuous as the former one. So this study provides us with a tentative scope of proper 
	$\lambda$ and $\mu$ for favorable activation performance, and such regularity implies that an applicable assignment of $\lambda$ and $\mu$ should be neither too large nor too 
	small.
	
	\par
\begin{figure}[!t]
		\centering
		\includegraphics[scale=0.25]{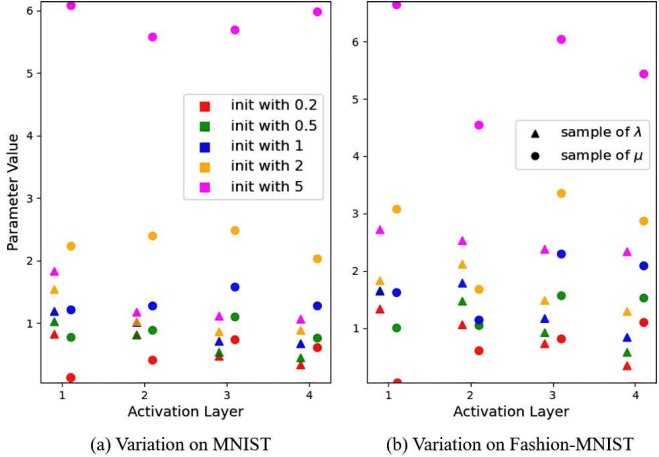}
		\caption{Parameters variation from different initialization on MNIST and Fashion-MNIST after learning. Triangular points are samples of $\lambda$ and circular ones are 
		samples of $\mu$. Initial values are distinguished by different colors.} 
		\label{Fig.4}
	\end{figure}

 Based on that, we narrow the value range and continue to investigate its preference when $\lambda$ and $\mu$ are capable of being learned dynamically from data. We keep 
	other training setups alike but increase the epoch to 30, and convert our activation function to learnable mode. Because there are just four activation layers in the official 
	structure of LeNet-5, we can easily record the variation tendency for each parameter after learning. More elaborately, we select five grades to start initialization for both 
	$\lambda$ and $\mu$, which are 0.2, 0.5, 1, 2, and 5. Finally, \autoref{Table2} and \autoref{Table3} give the updated results of two parameters after learning by 30 epochs, 
	and \autoref{Fig.4} shows the variation for each activation layer legibly.
	
	\par It can be found that there seems to be some kind of gravity in the vicinity of 1 that attracts $\lambda$ to update along, and those which initialized from both ends, 
	such as 0.2 and 5, tend to evolve towards the central zone. In contrast, $\mu$ appears not much interested in similar renewal but keeps itself fluctuating around the initial 
	value of every layer.
	
	\par From another perspective, this sort of discrepancy is sensible. Looking back on our PFPLUS formula in Eq.\eqref{PFPLUS-Piecewise}, $\lambda$ plays the role of amplitude 
	factor affecting both positive and negative domains. If it is too large, a gradient exploding problem will occur and the capability of resisting noise will be damaged. On the 
	contrary, if it is quite small, the trouble of gradient vanishing will impede the learning process. As for $\mu$, it only manipulates the speed of attenuation for negative 
	domain, and the activation level will decay to saturation sooner or later, which is just a matter of iteration time. So we know $\lambda$ has more impact on the quality of our 
	function compared to $\mu$, but that doesn't mean the latter one is neglectable since it likewise determines the optimal state of activation performance. 
	 
	 \par Therefore we see the phenomenon that $\lambda$ inclines to converge on 1 nearby after training, albeit its degree differs in the light of layer, whereas $\mu$ shows no 
	 apparent iteration tendency but oscillates around where it is initiated.
	
	\subsection{\textsf{Experiment on Kaggle's Dogs Vs Cats}}
	Kaggle's Dogs Vs Cats\footnote{URL: https://www.kaggle.com/biaiscience/dogs-vs-cats} is an adorable dataset released by Kaggle that contains 25K images in RGB format with 
	arbitrary sizes, and specimens of dogs and cats account for each half. We sample 10\% from every category randomly, which has 1250 examples respectively, and then distribute 
	them together to the validation set, so the rest part automatically constitutes the training set. 
	
	\par In the last subsection, we elementarily explore the influence of different specified values on learnable parameters, and this time we change the initialization scheme 
	based on previous work. To extend their variable range, we let the parameters obey some probability distributions rather than give them specific values, which may bring in 
	more flexibility and plasticity.
		
	\par Besides, we need a deeper network to verify the potentiality of our approach, so we choose the legendary masterpiece ResNet \cite{ResNet} with construction of 34 layers 
	as the framework, of which the activation units are replaced with our PFPLUS initialized in different random distributions. By reason of matching the input size that the 
	network requires, we adjust every image to $224 \times 224$ pixels before being fed in. The training epoch is 10 in this experiment with batch size 32, and the learning rate 
	is 0.001 but decays half every two epochs. Customarily, cross-entropy loss is the cost function and Adam is designated as the optimizer.
	
	\par For comparison, we take the constant distribution of 1 as the baseline, and select uniform distribution as well as normal distribution under conventional, Xavier 
	\cite{GradientVanishment}, and Kaiming \cite{PReLU} three ways to experiment.  In view of preceding speculation, we empirically test 25 combinations of various 
	distributions, contrasting their training diversification and identifying those good ones. The whole results are placed in the appendix and we excerpt the top-7 best groups
	to present in \autoref{Table4}.
	
	\begin{table}[!t]
		\centering
		\caption{The top-7 best initialization groups of learnable parameters obeying certain distribution, compared with baseline.}
		\small
		\begin{threeparttable}
		\begin{tabular*}{\linewidth}{@{\extracolsep{\fill}}cccc}
			\toprule
			$\lambda$ factor & $\mu$ factor & \makecell[c]{validation \\ loss} & \makecell[c]{validation \\ accuracy} \\
			\midrule
			1 			   & 1 			   & 0.2885 & 90.83\% \\
			U(0.5, 1.5)    & 1 			   & 0.1923 & 92.35\% \\
			Xavier Normal  & 1 			   & 0.2002 & 92.35\% \\
			Kaiming Normal & 1 			   & 0.1850 & 92.11\% \\
			1 			   & U(0.5, 1.5)   & 0.2687 & 92.35\% \\
			1 			   & N(1, $0.1^2$) & 0.2475 & 92.51\% \\
			N(1, $0.3^2$)  & N(1, $0.1^2$) & 0.2271 & 92.07\% \\
			Xavier Uniform & N(1, $0.1^2$) & 0.1725 & 92.99\% \\
			\bottomrule
		\end{tabular*}
		\begin{tablenotes}
		\footnotesize
		\item $\circledcirc$ $U(a,b)$ means uniform distribution with $a$ and $b$ being lower bound and upper bound. 
		\item $\circledcirc$ $N(\gamma,\sigma^2)$ means normal distribution with $\gamma$ and $\sigma^2$ being mean and variance. 
		\end{tablenotes}
		\end{threeparttable}
		\label{Table4}
	\end{table}
	
	\par Among all those collocations, we avoid applying Xavier and Kaiming variant distributions to $\mu$, since they will bring about negative initialization that violates the 
	restrictive definition of PFPLUS and fails to make target loss convergent, so Xavier and Kaiming ways are used for $\lambda$ only to make sure the coupled structure of 
	function won't be jeopardized.
	
	\par Judging from results of the experiment, we can see it may lead to a better effect if making the parameters subject to a suitable probability distribution, although 
	$\lambda$ and $\mu$ can be both initialized by 1 for the sake of efficiency.
	
	\begin{figure}[!t]
		\centering
		\includegraphics[scale=0.24]{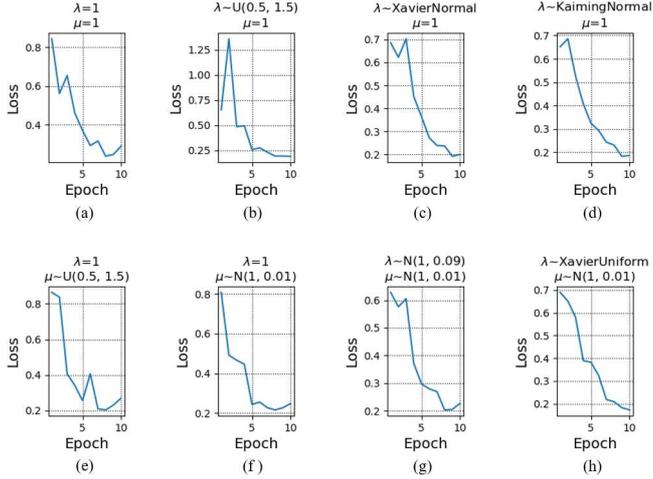}
		\caption{Curves of validation loss for the top-7 best initialization groups and the baseline subject to different distributions.} 
		\label{Fig.5}
	\end{figure}
	
	\begin{figure}[!t]
		\centering
		\includegraphics[scale=0.24]{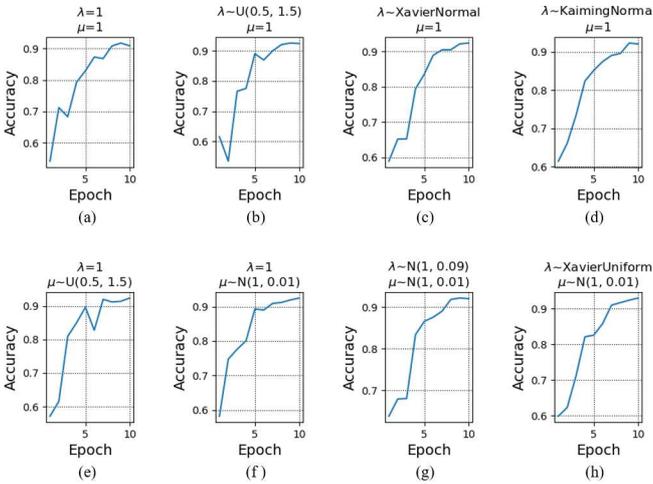}
		\caption{Curves of validation accuracy for the top-7 best initialization groups and the baseline subject to different distributions.} 
		\label{Fig.6}
	\end{figure}
	
	\par Checking on the entire 25 groups of experimental results, it is noticed that if adopting normal distribution separately, mean of 1 and standard deviation of 0.3 seem to 
	be more reasonable for $\lambda$, whereas standard deviation of 0.1 is better for $\mu$. About uniform distribution, boundaries of 0.5 and 1.5 are good choices for 
	both two parameters.
	
	\par However, given the overall statistics, we find groups of normal distribution usually achieve higher validation scores than those of uniform distribution, which can 
	be verified through composition proportion of the top 7 best ones as shown in \autoref{Table4}, dominated by Gaussian type distributions. Besides, under the same 
	circumstances, initialization of Xavier's version is more recommended to PFPLUS, since it is conducive to the activation that produces zero mean output, while Kaiming's way 
	principally aims at non-zero centered functions like ReLU.
	
	\begin{figure}[!t]
		\centering
		\includegraphics[scale=0.26]{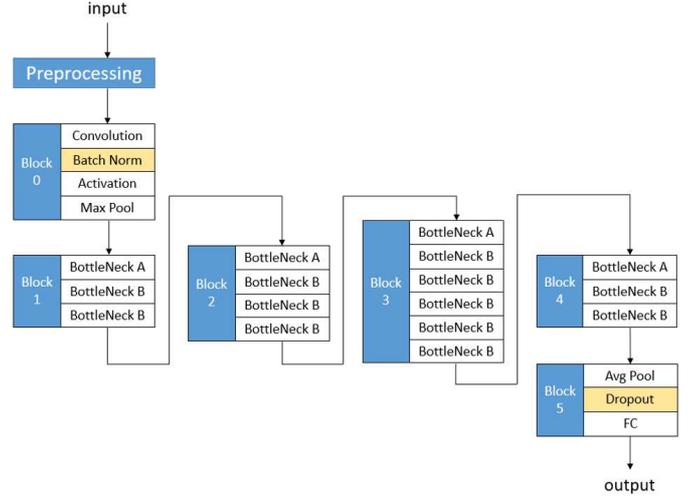}
		\caption{Architecture for traditional ResNet-50 with addition of dropout layer. Details in bottleneck A and B are shown below.} 
		\label{Fig.7}
	\end{figure}
	
	\begin{figure}[!t]
		\centering
		\includegraphics[scale=0.26]{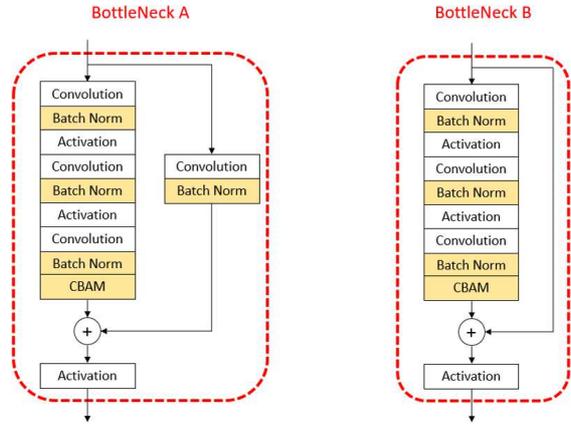}
		\caption{Structures of BottleNeck A and BottleNeck B in Fig. 7's construction, embedded with CBAM modules specially.} 
		\label{Fig.8}
	\end{figure}
	
	\par \autoref{Fig.5} and \autoref{Fig.6} illustrate the validation process and dynamic states of those effective distribution combinations in \autoref{Table4}, so we can 
	recognize that initialization of $\lambda$ obeying Xavier Uniform distribution while $\mu$ obeying normal distribution with 1 mean and 0.01 variance displays the steadiest 
	loss fall among 8 groups of curves, albeit set of $\lambda$\ following Kaiming normal distribution while $\mu$ following constant 1 distribution as well gains eligible 
	performance. As for accuracy, results of (c), (d) and (f), (g), (h) in \autoref{Fig.6} show comparatively stable ascending, where constant 1 distribution and Gaussian type 
	distributions involve for many times, so we think of them as preferences to our PFPLUS.
	
	\subsection{\textsf{Experiment on Intel Image Classification Dataset}}
	\par Intel has ever hosted an image classification challenge and released one scenery dataset\footnote{URL:https://www.kaggle.com/puneet6060/intel-image-classification}, 
	consisting of approximately 17K RGB images of size $150\times150$ pixels subordinate to 6 categories, which are buildings, forest, glacier, mountain, sea, and street. The 
	dataset splits in training part of 14K examples and test part of 3K, whose ratio is about $82\%:18\%$.
	
	\begin{table*}[!t]
		\centering
		\caption{Comparison of loss and accuracy for training Intel image classification dataset on ResNet-50 with multiple network configurations, grouped by two ways of 
		learning rate decay.}
		\small
		\begin{threeparttable}
		\begin{tabular*}{\linewidth}{@{\extracolsep{\fill}}ccccccccccc}
			\toprule
			\multicolumn{3}{c}{\multirow{2}*{Architecture Configuration}} & \multicolumn{4}{c}{Step Decay for lr} & 
			\multicolumn{4}{c}{Exponential Decay for lr} \\
			\cmidrule(lr){4-7} \cmidrule(lr){8-11}
			 & & & \multicolumn{2}{c}{Testing Loss} & \multicolumn{2}{c}{Testing Accuracy} & \multicolumn{2}{c}{Testing Loss} & \multicolumn{2}{c}{Testing Accuracy} \\
			\cmidrule{1-3} \cmidrule(lr){4-5} \cmidrule(lr){6-7} \cmidrule(lr){8-9} \cmidrule(lr){10-11}
			BatchNorm & CBAM & Dropout &ReLU&FPLUS &ReLU&FPLUS &ReLU&FPLUS &ReLU&FPLUS \\
			\midrule
			 & & & 0.674 & 0.484 & \color{cyan}75.24\% & \color{cyan}83.18\% & 1.098 & 0.822 & \color{cyan}55.39\% & \color{cyan}66.95\% \\
			$\checkmark$ & & & 0.682 & 0.544 & 87.28\% & 88.11\% & 0.678 & 0.669 & 86.21\% & 86.40\% \\
			 & $\checkmark$ & & 0.802 & 0.551 & \color{cyan}69.98\% & \color{cyan}80.19\% & 1.335 & 1.133 & \color{cyan}44.25\% & \color{cyan}55.75\% \\
			 & & $\checkmark$ & 0.722 & 0.499 & \color{cyan}72.57\% & \color{cyan}82.66\% & 1.123 & 0.835 & \color{cyan}53.19\% & \color{cyan}67.19\% \\
			$\checkmark$ & $\checkmark$ & & 0.563 & 0.571 & 86.90\% & 86.71\% & 0.607 & 0.609 & 85.79\% & 86.31\% \\
		    $\checkmark$ & & $\checkmark$ & 0.846 & 0.490 & 87.08\% & 87.44\% & 0.683 & 0.684 & 86.03\% & 86.51\% \\
			 & $\checkmark$ & $\checkmark$ & 0.846 & 0.617 & \color{cyan}67.47\% & \color{cyan}77.53\% & 1.306 & 1.151 & \color{cyan}46.14\% & \color{cyan}54.58\% \\
			$\checkmark$ & $\checkmark$ & $\checkmark$ & 0.586 & 0.496 & 87.50\% & 88.50\% & 0.651 & 0.666 & 85.71\% & 85.75\% \\
			\bottomrule
		\end{tabular*}
		\begin{tablenotes}
			\footnotesize
			\item $\circledcirc$ Accuracies in \textcolor{cyan}{cyan} color correspond to the results of trial groups that don't employ batch normalization layers.
			\item $\circledcirc$ Check mark $\checkmark$ means relevant module is added in the network.
		\end{tablenotes}
		\end{threeparttable}
		\label{Table5}
	\end{table*}
	
	\begin{figure*}[!t]
		\centering
		\includegraphics[scale=0.35]{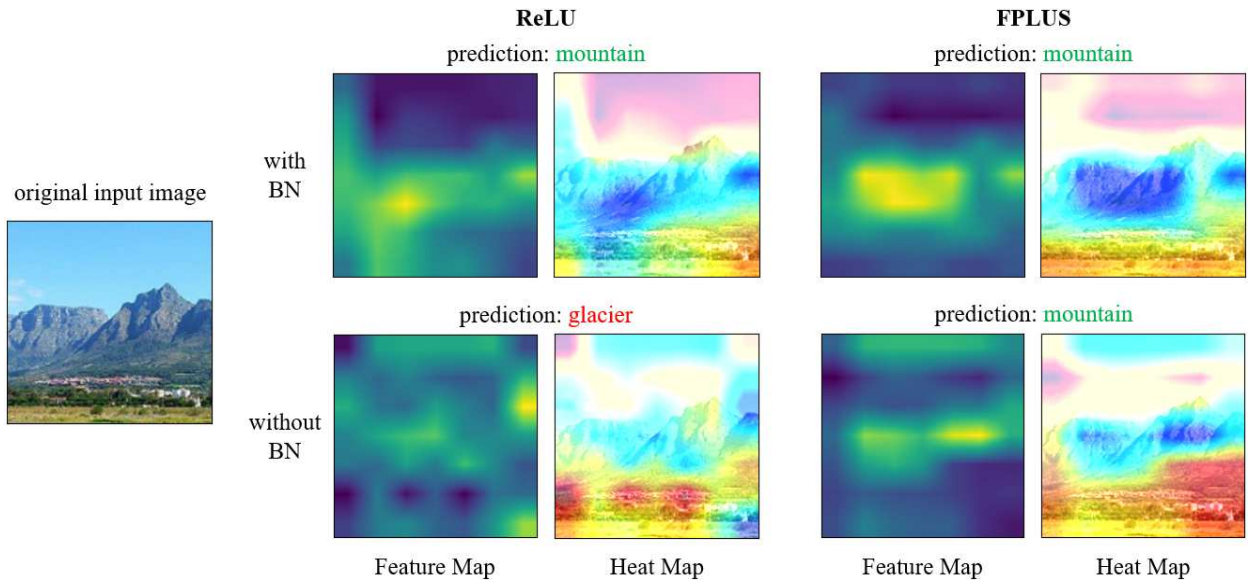}
		\caption{Visualization of feature maps and heat maps produced by model's final convolutional layer, under two circumstances whether batch normalization is employed or 
		not. Original input image as well as corresponding predictive results for ReLU and FPLUS are given in the figure to check.} 
		\label{Fig.9}
	\end{figure*}
	
	\par As is known to all, the iid condition is substantially a key to resolving the trouble that a training process becomes harder when the network is getting deeper, and this 
	kind of weakness is called bias shift effect \cite{ELU} or internal covariate shift \cite{BatchNorm}. Throughout all those reformative activation theories, batch normalization 
	provides such a solution driving inputs of each layer in the network to remain independently and identically distributed. 
	
	\par Taking ReLU \cite{Boltzmann_Machines,ReLU} as an example, it directly prunes the activation of negative domain for every hidden neuron, making the entire distribution of 
	input data deviates to limit region of the mapping function, so that gradient of lower level in the network probably disappears during backward propagation, and batch 
	normalization exactly serves for dragging it back to a normal distribution with 0 mean and 1 variance.
	
	\par Nevertheless, unlike ReLU's bottom truncation, our function FPLUS retains mild activation for negative zone in a gradually saturated way, so as to counteract the shift 
	influence in some degree. Therefore it means ReLU pretty much relies on batch normalization theoretically, while FPLUS remains more tolerable even if without it, affected by 
	less impact than ReLU.
	
	\par In order to validate the aforementioned hypothesis, we conduct an ablation study in this subsection, mainly researching the influence of batch normalization to 
	non-linear activation, so we compare the responses of ReLU and our FPLUS under two situations whether batch normalization is utilized or not. 
	
	\par As is shown in \autoref{Fig.7} and \autoref{Fig.8}, we opt for ResNet-50 \cite{ResNet} to be the framework, but extra introduce the regularization mechanism dropout 
	\cite{Dropout} and attention module CBAM \cite{CBAM}, to explore their mutual effect. Besides, we simultaneously try two attenuation styles for learning rate, including step 
	decay and exponential decay, to investigate which one suits more.
	
	\par We train the model for 30 epochs with batch size 64 and choose cross-entropy loss like before. The optimizer is SGD with 0.9 being momentum, and the learning rate starts 
	from 0.001, multiplied by 0.1 every 10 epochs in the event of step decay, and making 0.98 the base if it employs exponential decay.
	
	\par From the results compared in \autoref{Table5} we know, on the condition that applying the same decay mode of learning rate, settings with batch normalization are able 
	to produce quite close outcomes for ReLU and FPLUS, while trials with step decay achieve slightly better grades than exponential decay if using identical activation function. 
	On the contrary, if banning batch normalization, both decaying ways for learning rate will generate fallen performance, which is greatly apparent for exponential attenuation. 
	However, our method FPLUS appears more solid and reliable, because it obtains considerably better results than ReLU when they both confront the lack of batch normalization, 
	and this can be seen from the contrast of entries with cyan color in \autoref{Table5}.
	
	\par According to the ablation study, separate utilization of dropout or CBAM without batch norm seems to degrade the quality of activation, and such fact manifests the 
	important position of batch norm, which is deemed to lay the foundation for the other two if we want improved performance through combining them.
	
	\par \autoref{Fig.9} shows the different responses of ReLU and FPLUS under two contrastive circumstances whether batch normalization is used or not, and both feature maps 
	and heat maps are given to intuitively illustrate their reactions generated by distinct activation layers.
	
	\par When the network enables batch normalization, models embedded respectively with ReLU and FPLUS can both output correct prediction, but feature map of ReLU is much 
	more dispersive than that of FPLUS. For heat maps, the response area of FPLUS precisely lies on the key object, whereas ReLU's does not. 
	
	\par On the other hand, when batch normalization is disabled, models activated by ReLU and FPLUS all produce degenerative feature maps, among which ReLU's is more 
	ambiguous while that of FPLUS is a little bit better. With regard to heat maps, except for their common flaws at the top, the one belonging to FPLUS concentrates relatively 
	more on valid targets, yet ReLU's is interfered with by something else in the background. Besides, it's worth noting that classifier of ReLU misjudges the image category, 
	whereas prediction of FPLUS still hits the right answer.
	
	\par All things considered, the capacity of batch normalization is absolutely pivotal, on which ReLU relies very much, but our method comparatively shows less dependency and 
	frangibility, notwithstanding it would be better for sure if batch normalization can be exploited at the same time.
	
	\subsection{\textsf{Experiments on CIFAR-10}}	
	\begin{table}[!t]
		\centering
		\caption{Networks trained on CIFAR-10 and their parameter quantities as well as calculation overhead.}
		\small
		\begin{tabular*}{\linewidth}{@{\extracolsep{\fill}}crc}
			\toprule
			Network 	 					& \# Params  &  FLOPs \\
			\midrule
			SqueezeNet \cite{SqueezeNet}    & 0.735M 	 & 0.054G \\
			NASNet \cite{NASNet}			& 4.239M 	 & 0.673G \\
			ResNet-50 \cite{ResNet}			& 23.521M 	 & 1.305G \\
			InceptionV4 \cite{InceptionV4}  & 41.158M 	 & 7.521G \\
			\bottomrule
		\end{tabular*}
		\label{Table6}
	\end{table}
	
	\begin{table}[!t]
		\centering
		\caption{Loss on validation set of CIFAR-10 for each network with different activation functions.}
		\small
		\begin{threeparttable}
			\begin{tabular}{ccccc}
				\toprule
				\thead{Activation}   & \thead{SqueezeNet} & \thead{NASNet} & \thead{ResNet-50} & \thead{InceptionV4} \\
				\midrule
				ReLU   & 0.350 & 0.309 & 0.335 & 0.442 \\
				LReLU  & 0.345 & 0.307 & 0.322 & 0.412 \\
				PReLU  & 0.365 & 0.366 & 0.370 & 0.305 \\
				ELU    & 0.360 & \color{red}0.280 & 0.259 & 0.234 \\
				SELU   & 0.447 & 0.307 & 0.295 & 0.307 \\
				FPLUS  & 0.349 & 0.287 & 0.255 & \color{red}0.220 \\
				PFPLUS & \color{red}0.344 & 0.323 & \color{red}0.251 & 0.230 \\
				\bottomrule
			\end{tabular}
			\begin{tablenotes}
				\footnotesize
				\item $\circledcirc$ Entries in \textcolor{red}{red} color indicate the best results for each network under validation set. 
			\end{tablenotes}
		\end{threeparttable}
		\label{Table7}
	\end{table}
	
	\begin{table}[!ht]
		\centering
		\caption{Accuracy rate (\%) on validation set of CIFAR-10 for each network with different activation functions.}
		\small
		\begin{threeparttable}
		\begin{tabular}{ccccc}
			\toprule
			\thead{Activation}   & \thead{SqueezeNet} & \thead{NASNet} & \thead{ResNet-50} & \thead{InceptionV4} \\
			\midrule
			ReLU   & 88.67 & 91.55 & 91.94 & 87.95 \\
			LReLU  & 88.63 & 91.61 & 92.20 & 88.18 \\
			PReLU  & 88.91 & 91.52 & 92.09 & 92.95 \\
			ELU    & 87.82 & \color{magenta}91.70 & 92.71 & 92.59 \\
			SELU   & 84.83 & 90.35 & 91.10  & 90.53 \\
			FPLUS  & 88.70 & 91.46 & 92.75 & \color{magenta}93.47 \\
			PFPLUS & \color{magenta}89.09 & 91.62 & \color{magenta}93.13 & 93.41 \\
			\bottomrule
		\end{tabular}
		\begin{tablenotes}
			\footnotesize
			\item $\circledcirc$ Entries in \textcolor{magenta}{magenta} color indicate the best results for each network under validation set. 
		\end{tablenotes}
		\end{threeparttable}
		\label{Table8}
	\end{table}

	\par Dataset of CIFAR-10\footnote{URL: http://www.cs.toronto.edu/~kriz/cifar.html} comprises 60K samples with image size of $32\times32$ pixels in RGB format. It splits into a 
	training set of 50K instances and a validation set of 10K. Training set and validation set have covered each category proportionally for both.
	
	\par Having discussed the intrinsic values of our formulation in previous experiments, now we plan to explore its compatibility with various network architectures and 
	compare it with other more typical activation functions.
	
	\par Firstly, we select 4 kinds of networks from lightweight to heavyweight as the foundation frameworks to train on CIFAR-10, and their numbers of parameters as well as the 
	amount of calculation are listed in \autoref{Table6}. Varieties of activation ways also include ReLU's improvers LReLU \cite{Leaky_ReLU} and PReLU \cite{PReLU}, as well as 
	ELU \cite{ELU} and its variant SELU \cite{SELU}.
	
	\begin{figure}[!t]
		\centering
		\includegraphics[scale=0.29]{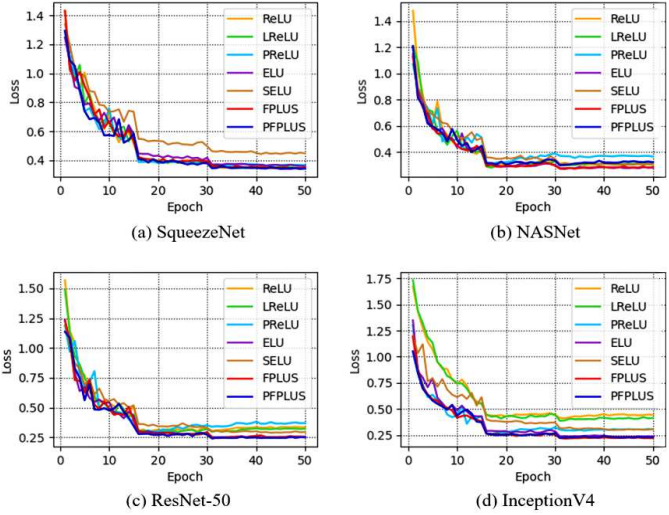}
		\caption{Curves of validating loss for each network applying diverse activation methods.} 
		\label{Fig.10}
	\end{figure}
	
	\begin{figure}[!t]
		\centering
		\includegraphics[scale=0.29]{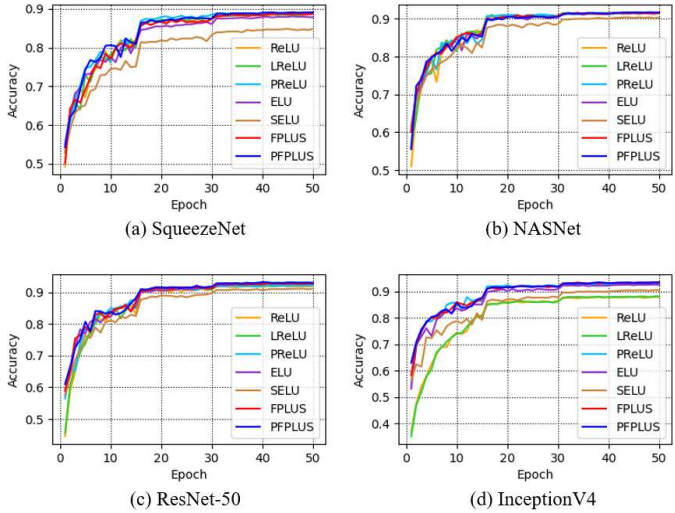}
		\caption{Curves of validating accuracy for each network applying diverse activation methods.} 
		\label{Fig.11}
	\end{figure}
	
	\par For 4 groups of training, setups of hyperparameters keep the same, and we still use SGD with 0.9 momentum to optimize cross-entropy loss. The iterative epoch is 50 
	with batch size being 128, and the learning rate starts by 0.01, multiplied by 0.2 every time at 15th, 30th, 40th epoch for decay, which represent the checkpoints of $30\%$, 
	$60\%$ and $80\%$ in training process. For the purpose of data augmentation, we take some measures to transform the training set, such as random cropping, flipping, and 
	rotation.

	\par Loss and accuracy on validation set for each network with distinct activation ways are presented in \autoref{Table7} and \autoref{Table8}, in which colored items stand 
	for the best results of corresponding networks. We can see our method FPLUS and PFPLUS clearly outperform the others on ResNet-50 and InceptionV4, while that is not so evident 
	on SqueezeNet or NASNet.
	
	\par In addition, we completely record their validating processes of loss and accuracy, as shown in \autoref{Fig.10} and \autoref{Fig.11}. It can be noticed that for 
	heavyweight models like ResNet-50 and InceptionV4, our activation units display swifter loss descending as well as more prominent accuracy rising that enable them to be 
	distinguished from others.
	
	\par As a consequence, we think our activation function is an effective and reliable approach that is capable of standing the trial, even if confronting a large workload of 
	high intensity such as ImageNet.
	
	\par Anyway, in comprehensive consideration of the performance on CIFAR-10, we hold that our proposed method shows comparable competitiveness and invariable stability, which 
	make for the promotion of accommodating to as many integrated networks as possible.
 
	\section{Conclusion} \label{Sec6}
	\par Taking inspiration from conceptual bionics and inverse operation, this paper proposes a novel and subtle activation method, which involves mathematical first power and 
	switch-like sign function. We theoretically derive the formula under certain specified prior knowledge, and get a concise expression form that owns two adjustable parameters. 
	Either of them regulates the shape of our function, while they can be fixed or learnable. We name our parametric formulation PFPLUS, and FPLUS is one particular case. 
	Furthermore, we investigate the influence of initialization variety for trainable parameters, and find out the limited interval near 1 is a preferable option with 
	interpretability to achieve good performance. Meanwhile, we conduct an ablation study to demonstrate our activation unit is not that vulnerable to the lack of batch 
	normalization, showing less dependency but more robustness. Besides, when compared with other typical activation approaches in the same task, our new way manifests not only 
	appreciable competence but also consistent stability, which provides considerable advantages to its undeniable competitiveness. On the whole, we hold the opinion that our 
	work to some extent enriches the diversity of activation theories.
	
	\bibliography{references}
	
	\appendix 
	\section{Details for parameter initialization of diverse distributions in Sec. 5.2} 
		\setcounter{table}{0}
		\begin{table}[!h]
			\centering
			\caption{25 groups of different initialization for learnable parameters obeying certain distribution, compared with baseline.}
			\footnotesize
			\begin{tabular}{ccccc}
				\toprule
				No. & \makecell{$\lambda$ \\ Factor} & \makecell{$\mu$ \\ Factor} & \thead{\makecell{Validation \\ Loss}} & \thead{\makecell{Validation \\ Accuracy}} \\
				\midrule
				1     & 1 				& 1 			& 0.2885 & 90.83\% \\
				2     & U(0.5, 1.5) 	& 1 			& 0.1923 & 92.35\% \\
				3     & U(0.9, 1.1) 	& 1 			& 0.2577 & 91.31\% \\
				4     & N(1, $0.3^2$) 	& 1 			& 0.2573 & 91.39\% \\
				5     & N(1, $0.1^2$) 	& 1 			& 0.2944 & 88.70\% \\
				6     & XavierUniform 	& 1 			& 0.2012 & 91.23\% \\
				7     & XavierNormal 	& 1 			& 0.2002 & 92.35\% \\
				8     & KaimingUniform  & 1 			& 0.2117 & 91.03\% \\
				9     & KaimingNormal 	& 1 			& 0.1850 & 92.11\% \\
				10    & 1 				& U(0.5, 1.5)   & 0.2687 & 92.35\% \\
				11    & 1 				& U(0.9, 1.1)   & 0.2535 & 90.75\% \\
				12    & 1 				& N(1, $0.3^2$) & 0.2894 & 90.83\% \\
				13    & 1 				& N(1, $0.1^2$) & 0.2475 & 92.51\% \\
				14    & U(0.5, 1.5) 	& U(0.5, 1.5)   & 0.2450 & 91.67\% \\
				15    & U(0.5, 1.5) 	& N(1, $0.1^2$) & 0.2398 & 91.75\% \\
				16    & N(1, $0.3^2$) 	& N(1, $0.1^2$) & 0.2271 & 92.07\% \\
				17    & N(1, $0.3^2$) 	& U(0.5, 1.5)   & 0.2333 & 91.71\% \\
				18    & XavierNormal 	& U(0.5, 1.5)   & 0.2141 & 91.55\% \\
				19    & XavierNormal 	& N(1, $0.1^2$) & 0.2080 & 91.11\% \\
				20    & XavierUniform 	& U(0.5, 1.5)   & 0.2298 & 90.71\% \\
				21    & XavierUniform 	& N(1, $0.1^2$) & 0.1725 & 92.99\% \\
				22    & KaimingNormal 	& U(0.5, 1.5)   & 0.2002 & 91.87\% \\
				23    & KaimingNormal 	& N(1, $0.1^2$) & 0.2506 & 90.26\% \\
				24    & KaimingUniform 	& U(0.5, 1.5)   & 0.2670 & 89.46\% \\
				25    & KaimingUniform 	& N(1, $0.1^2$) & 0.3199 & 88.38\% \\
				\bottomrule
			\end{tabular}
			\label{Appendix Table}
		\end{table}

\end{document}